\documentclass{article}
\usepackage{spconf,amsmath,graphicx}
\usepackage{cite,bm}
\usepackage{xcolor}
\usepackage{booktabs}

\title{SpeeChain: a speech toolkit for large-scale machine speech chain}
\name{Heli Qi$^1$, 
Sashi Novitasari$^{1*}$, 
Andros Tjandra$^{1\dagger}$, 
Sakriani Sakti$^2$, 
Satoshi Nakamura$^1$}
\address{
  $^1$Nara Institute of Science and Technology, Japan\\
  $^2$Japan Advanced Institute of Science and Technology, Japan}

\begin{document}
%\ninept
%
\maketitle
\renewcommand{\thefootnote}{\fnsymbol{footnote}}
\footnotetext[1]{Sashi Novitasari is currently with IBM Tokyo Research Lab.}
\footnotetext[2]{Andros Tjandra is currently with Meta AI.}
\begin{abstract}
% toolkit overview
This paper introduces SpeeChain, an open-source Pytorch-based toolkit designed to develop the machine speech chain for large-scale use. 
% This first release focuses on applying the machine speech chain to TTS data augmentation for ASR, i.e. TTS$\to$\\ASR chain. 
This first release focuses on the TTS$\to$ASR chain, a core component of the machine speech chain, that refers to the TTS data augmentation by unspoken text for ASR.
% Multi-GPU batch-level model inference, multi-dataloader batch generation
To build an efficient pipeline for the large-scale TTS$\to$ASR chain, we implement easy-to-use multi-GPU batch-level model inference, multi-dataloader batch generation, and on-the-fly data selection techniques.
% LibriSpeech-100&360，in-domain semi-supervised ASR
% In this paper, we extend our previous experiments of TTS$\to$ASR chain to the large-scale LibriSpeech corpus where we only use \emph{train\_clean\_100} as labeled data for training ASR and TTS base models. 
% We show detailed experimental analyses of how to improve the ASR performance by TTS synthetic speech. 
In this paper, we first explain the overall procedure of the TTS$\to$ASR chain and the difficulties of each step. 
Then, we present a detailed ablation study on different types of unlabeled data, data filtering thresholds, batch composition, and real-synthetic data ratios.
% With thorough experiments on \emph{train\_clean\_460} of LibriSpeech, we demonstrate that our TTS$\to$ASR chain can significantly improve WER by a large amount of unspoken text. 
% We will release our toolkit upon publication.
Our experimental results on \emph{train\_clean\_460} of LibriSpeech demonstrate that our TTS$\to$ASR chain can significantly improve WER in a semi-supervised setting.\footnote[3]{We will release our toolkit upon publication.}

% \jinming{
% The toolkit is based on Pytorch and supports easy-to-use multi-GPU training and inference.  
% This first release focuses on TTS$\to$ASR chain that executes TTS data augmentation for ASR, a core component of speech chain. 
% In the first release, we focus on the usage of the machine speech chain on TTS data augmentation for ASR, i.e. TTS$\to$ASR chain.
% Multi-GPU batch-level model inference, multi-dataloader batch generation
% To build a fast and stable framework for large-scale TTS$\to$ASR chain, we implement easy-to-use multi-GPU batch-level model inference and multi-dataloader batch generation techniques in our toolkit.
% LibriSpeech-100&360，in-domain semi-supervised ASR
% In this paper, we extend our previous experiments of TTS$\to$ASR chain to the large-scale LibriSpeech corpus where we only use \emph{train\_clean\_100} as labeled data for training supervised ASR and TTS base models. 
% We show detailed experimental analyses of how to improve the ASR performance by TTS synthetic speech. 
% With thorough experiments, we demonstrate that TTS$\to$ASR chain can significantly improve ASR performance with a large amount of unspoken text.\footnote{We will release our toolkit and models upon publication. }% 模型开源和数据开源是个很大的亮点！！！！！
% }

\end{abstract}
\section{Introduction}
\label{sec:intro}
% 研究背景
% 端到端ASR与TTS，深度神经网络
End-to-end (E2E) automatic speech recognition (ASR) \cite{chorowski2015attention,chan2016listen,dong2018speech,hannun2019sequence} and text-to-speech synthesis (TTS) \cite{shen2018natural,li2019neural,chen2020multispeech} have achieved great success due to recent advances in deep learning. 
% 简短描述一下E2E ASR和TTS模型功能，提出他们在输入输出部分具有相似性
% E2E ASR models directly convert an input speech signal to the corresponding transcript while E2E TTS models generate the speech signal based on an input text. 
% ASR and TTS models are symmetric with each other since they share the same type of training data but have opposite input and output.
ASR and TTS are considered symmetric to each other, as they share the same type of training data with flipped inputs and outputs. 
% Despite their significant similarity, ASR and TTS research progressed more or less independently for a long time in the past without exerting much influence on each other.
Despite the close relationship, research on both areas progressed more or less independently until \cite{tjandra2017listening,tjandra2020machine} proposed the machine speech chain. 
Research on E2E ASR-TTS has received more attention ever since \cite{ren2019almost,wang2020improving}.

% \jinming{ 
% [Below is to replace the above two paragraphs] 
% Automatic speech recognition (ASR) and text-to-speech synthesis (TTS) are considered symmetric to each other, as they share the same type of training data with flipped inputs and outputs.
% Despite the close relationship, research on both areas mostly progressed independently until machine speech chain \cite{tjandra2017listening,tjandra2020machine} was proposed. 
% Research on E2E ASR-TTS has received more attention ever since \cite{citation needed}.} % 后续引文

% 相关工作
% 近年来在speech chain的概念提出来之后，ASR-TTS的研究逐渐变得多了起来
% In recent years, ASR-TTS joint research has received more and more attention after the machine speech chain \cite{tjandra2017listening,tjandra2020machine} was proposed.
% 最开始是single-speaker数据集上做实验，无需speaker embedding
% In the original machine speech chain \cite{tjandra2017listening}, ASR is trained on synthetic speech produced by TTS using unspoken text (TTS$\to$ASR chain). 
Machine speech chain is a closed-loop architecture based on deep learning that connects E2E ASR and TTS models.
One core application of the machine speech chain is the TTS$\to$ASR chain, where E2E ASR models are trained on synthetic speech produced by E2E TTS models using unspoken text.
% 然后过渡到multi-speaker，需要speaker embedding的介入。
% The initial TTS$\to$ASR experiment on single-speaker datasets \cite{tjandra2017listening} was successful, which has led to an interest in training E2E ASR models with synthetic speech produced by multi-speaker TTS models.
The initial experiments of the TTS$\to$ASR chain on single-speaker \cite{tjandra2017listening} and multi-speaker datasets \cite{tjandra2018machine} were successful, which has led to an interest in training E2E ASR models with synthetic speech produced by multi-speaker TTS models in the community \cite{rossenbach2021comparing,li2018training,rossenbach2020generating,rosenberg2019speech,ueno2019multi}.
For multi-speaker TTS, a speaker embedding model is introduced to help TTS systems capture various speaker identities and generate synthetic speech with different speaking styles. 
% \cite{tjandra2018machine} achieved the one-shot speaker adaptation that enabled TTS models to handle unseen speakers. 
Moreover, \cite{rosenberg2019speech} proposed two additional embedding encoders to extract local and global utterance embeddings for better TTS fidelity. 
Apart from speaker diversity, \cite{rosenberg2019speech,ueno2019multi,yue2021exploring,baskar2021eat} provided TTS systems with unlabeled out-of-domain text to help ASR models overcome linguistic mismatch. 

% \jinming{ 
% One core aspect in machine speech chain is TTS$\to$ASR chain where E2E ASR models are trained with synthetic speech produced by E2E TTS models using unspoken text.
% The initial TTS$\to$ASR experiment on single-speaker datasets \cite{tjandra2017listening} was successful, which leads to an interest in training E2E ASR models with synthetic speech produced by multi-speaker TTS models. 
% \cite{rossenbach2021comparing,li2018training,tjandra2018machine,rossenbach2020generating,rosenberg2019speech,ueno2019multi} introduced a speaker embedding model that helped TTS systems capture various speaker identities and generate synthetic speech with different speaking styles. 
% \cite{tjandra2018machine} achieved one-shot speaker adaptation that enabled TTS models to handle unseen speakers. 
% \cite{rosenberg2019speech} introduced two additional embedding encoders to extract local and global utterance embeddings for better TTS fidelity. 
% Apart from speaker diversity, \cite{rosenberg2019speech,ueno2019multi,yue2021exploring,baskar2021eat} provided TTS systems with unlabeled out-of-domain text to help ASR models overcome domain mismatch.\footnote{Note for Heli: please make sure all papers cited in this paragraph worked on joint ASR-TTS.}
% }

% 工具包类研究存在问题
% ASR-TTS相关的工具包汇总
In the speech processing community, many excellent open-source toolkits \cite{watanabe2018espnet,hayashi2020espnet,ott2019fairseq,wang2021fairseq,ravanelli2021speechbrain,zhang2022paddlespeech} are developed to support ASR and TTS models, and they have achieved comparable performance to the state-of-the-art. 
However, most of these all-in-one toolkits develop the ASR and TTS components independently with little linkage. 
% gap：没有一个统一的toolkit来进行半监督ASR以及ASR-TTS joint research
% To bridge the gap between ASR and TTS, our SpeeChain toolkit offers researchers and developers an easy-to-use pipeline of TTS$\to$ASR chain including fast model inference by multiple GPUs, uniform storage structure for real and synthetic data, and on-the-fly data filtering and loading services. 
% To bridge the gap between ASR and TTS, we present SpeeChain, a toolkit that offers researchers and developers an easy-to-use pipeline of the TTS$\to$ASR chain. 
To bridge the gap and establish the connection between ASR and TTS, we present SpeeChain, a toolkit that offers researchers and developers an efficient pipeline of the TTS$\to$ASR chain. 
% For an efficient pipeline of the large-scale TTS$\to$ASR chain, our toolkit supports various modular functions with flexibility such as fast model inference by multiple GPUs, uniform storage structure for real and synthetic data, and on-the-fly data filtering and loading. 
% Also, we provide intensively-tuned hyper-parameters for our ASR and TTS models to facilitate research on joint ASR-TTS.
SpeeChain supports various modular functions, such as fast model inference by multiple GPUs, uniform storage structure for real and synthetic data, and on-the-fly data filtering and loading. 
Furthermore, we provide intensively-tuned hyper-parameters for our ASR and TTS models to facilitate research on joint ASR-TTS.

% \jinming{
% % In the speech processing community, many great open-source toolkits \cite{watanabe2018espnet,hayashi2020espnet,ott2019fairseq,wang2021fairseq,ravanelli2021speechbrain,zhang2022paddlespeech} that support ASR and TTS achieves comparable performance to state-of-the-art. 
% However, these toolkits implement the ASR and TTS mostly independently. 
% % For an efficient implmentation of TTS$\to$ASR chain, the key point is the generation, storage, and usage of synthetic speech. 
% To bridge the gap and link ASR and TTS, we present SpeeChain, a toolkit that offers researchers and developers an easy-to-use pipeline of TTS$\to$ASR chain. It supports various modular functions with flexibility, while hyper-parameters for ASR and TTS models are intensively tuned. We hope this toolkit to facilitate research on joint ASR-TTS.
% % including fast model inference by multiple GPUs, uniform storage structure for real and synthetic data, and on-the-fly data filtering and loading services. 
% }

% 简要介绍本文工具包用途以及本文所作的实验
The rest of the paper is organized as follows:
Section \ref{sec:TTS2ASR_chain} describes the procedure of the TTS$\to$ASR chain. 
Section \ref{sec:toolkit} explains the technical solutions of our toolkit to the TTS$\to$ASR chain.
Section \ref{sec:exp} presents our experimental results.
In Section \ref{sec:conclusion}, we conclude this work and introduce our future work.

% 这个sec放到引言后面，然后toolkit overview放到这里当第三个sec
\section{TTS Data Augmentation for ASR\\(TTS$\to$ASR Chain)}
\label{sec:TTS2ASR_chain}
\begin{figure}[t]
  \centering
  \includegraphics[width=\linewidth]{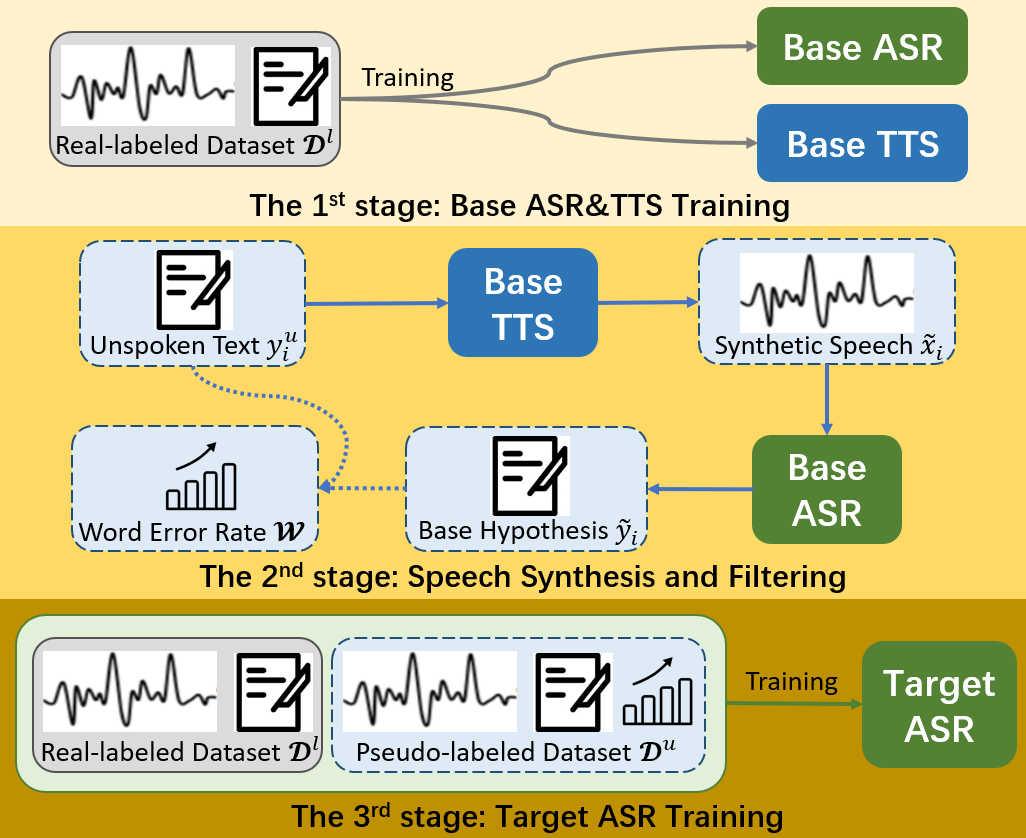}
  \caption{The overview of the TTS$\to$ASR chain.}
  \label{fig:semi_asr}
\end{figure}

% 具体定义一下本工具包中所指代的半监督ASR的含义
The procedure of the TTS$\to$ASR chain is presented in Fig. \ref{fig:semi_asr}.

\subsection{Base ASR\&TTS training}
% 第一阶段：基模型训练。这个stage的难点在于如何在少量有标签数据上训练一个尽可能好性能的模型，因为基模型的性能很大程度决定了后续目标模型可以达到的天花板
% long warming-up, smaller learning rates, train for more epochs, regularization, less model parameters
In the beginning, base ASR and TTS models are trained on the same labeled dataset. 
The performance of the base models largely determines the quality of synthetic speech and its WER evaluation in the next stage. 
However, due to the incomplete distribution of the limited labeled data, it's hard to train a base model that performs reasonably well, especially for multi-speaker TTS. 
Commonly-used strategies include elongating the warm-up stage, lowering the learning rate, raising dropout rates \cite{srivastava2014dropout}, shrinking the model parameters, etc.

\subsection{Speech synthesis and filtering}
% 第二阶段：伪数据生成。这个stage的难点在于受基模型性能的影响，伪数据中会存在很多errors，如何更正这些error或者去过滤掉低质量伪数据决定了后续目标模型可以达到的天花板
% language model calibration, lexicon calibration, rule-based manual correction, confidence & uncertainty filtering, 
In this stage, given an unlabeled dataset $\mathcal{D}^{u}=\{y^u_1,\dots, y^u_{|\mathcal{D}^{u}|}\}$, the base TTS model $\theta^{base}_{TTS}$ generates synthetic speech $\Tilde{x}_i$ for unspoken text $y^u_i$ ($i \in \{1,\dots,|\mathcal{D}^{u}|\}$) as
\begin{equation}
  \Tilde{x}_i = \mathop{\arg\max}\limits_{z} \prod_{t=1}^T{p(z_t|\Tilde{x}_{i,0:t-1},y_i^u;\theta_{TTS}^{base})}.
\label{eq1}
\end{equation}

The synthetic speech $\Tilde{x}_i$ is combined with unspoken text $y^u_i$ to form the pseudo-labeled dataset $\mathcal{D}^{u}=\{(\Tilde{x}_1,y^u_1),\dots,\\(\Tilde{x}_{|\mathcal{D}^{u}|},y^u_{|\mathcal{D}^{u}|})\}$.
However, despite the considerable effort in training the base TTS model, there may still be some errors in the synthetic speech (e.g., mispronunciations, silences, repeating phrases) because of the limitation on the amount of available labeled data. 
Data filtering is usually done to reduce the errors in the synthetic speech before the final stage.  
The TTS$\to$ASR chain enables us to filter the synthetic speech via the base word error rate (WER) $\mathcal{W}^{base}$ as
\begin{equation}
  \Tilde{y}_i = \mathop{\arg\max}\limits_{z} \prod_{t=1}^T{p(z_t|\Tilde{y}_{i,0:t-1},\Tilde{x}_i;\theta_{ASR}^{base})},
\label{eq2}
\end{equation}
\begin{equation}
  \mathcal{W}^{base}_i = \frac{edit\_distance(\Tilde{y}_i,y_i^u)}{|y_i^u|}
\label{eq3}
\end{equation}
where $\Tilde{y}_i$ is the hypothesis made by the base ASR model for $\Tilde{x}_i$, $edit\_distance()$ is the function that measures the minimum edit distance between the word sequences of $\Tilde{y}_i$ and $y_i^u$, and $|y_i^u|$ is the number of words in $y_i^u$.

\subsection{Target ASR training}
% 第三阶段：目标模型训练。这个stage的难点在于如何利用pseudo data训练模型，因为error的问题，pseudo data计算得到的gradient可能会存在很多噪音。
In the final stage, the synthetic speech whose $\mathcal{W}^{base}$ is larger than the given threshold will be first filtered out from the pseudo-labeled dataset $\mathcal{D}^u$. 
Then, the union of the real-labeled dataset $\mathcal{D}^{l}=\{(x^l_1, y^l_1),\dots,(x^l_{|\mathcal{D}^{l}|}, y^l_{|\mathcal{D}^{l}|})\}$ and $\mathcal{D}^{u}$ gives us an enlarged one $\mathcal{D}=\mathcal{D}^{l} \cup \mathcal{D}^{u}$ where we could train our target ASR model $\theta^{target}_{ASR}$. 
% The target ASR model could be initialized either randomly (i.e. training from scratch) or from the parameters of the base ASR model (i.e. retraining).
% basic version of ASR self-training and TTS data augmentation, teacher-forcing by pseudo data, use math equations here
During training, a single batch $\mathcal{B}=\mathcal{B}^l \cup \mathcal{B}^u$ is composed of real-labeled data $\mathcal{B}^l=\{(x^l_1, y^l_1),\dots,(x^l_{|\mathcal{B}^{l}|}, y^l_{|\mathcal{B}^{l}|})\}$ and pseudo-labeled data $\mathcal{B}^u=\{(\Tilde{x}_1, y^u_1),\dots,(\Tilde{x}_{|\mathcal{B}^{u}|}, y^u_{|\mathcal{B}^{u}|})\}$ proportionally fetched from $\mathcal{D}^{l}$ and $\mathcal{D}^{u}$. 
Two training losses $\mathcal{L}^l$ and $\mathcal{L}^u$ are calculated on $\mathcal{B}^l$ and $\mathcal{B}^u$ respectively as
\begin{equation}
  \mathcal{L}^{l} = - \frac{1}{|\mathcal{B}^l|} \sum_{i=1}^{|\mathcal{B}^l|} \sum_{t=1}^T \log{p(y^l_{i,t}|y^l_{i,0:t-1}, x^l_i; \theta^{target}_{ASR})},
\label{eq4}
\end{equation}
\begin{equation}
  \mathcal{L}^{u} = - \frac{1}{|\mathcal{B}^u|} \sum_{i=1}^{|\mathcal{B}^u|} \sum_{t=1}^T \log{p(y^u_{i,t}|y^u_{i,0:t-1}, \Tilde{x}_i; \theta^{target}_{ASR})},
\label{eq5}
\end{equation}
where both speech $x_i$ and prefix tokens $y_{0:t-1}$ are used as input based on the teacher-forcing technique. Finally, the target ASR model is optimized by the overall loss $\mathcal{L} = (1-\lambda) \mathcal{L}^l + \lambda \mathcal{L}^u$ controlled by a manually-set weight $\lambda$.

% 为了空间可以删除，more advanced training schemes
% Beyond the basic semi-supervised algorithm above, there are many more advanced and well-designed algorithms. 
% For example, \cite{kahn2020self} adopts model ensemble strategy to improve the quality of pseudo text, \cite{park2020improved,xu2020iterative} take the target ASR as the new base model to recursively conduct self-training, \cite{likhomanenko2020slimipl,higuchi2022momentum} simultaneously carry out the last two stages online, and so on.
% iterative
% loss weight adjustment (reduce the confidence degree of pseudo data)
% degree of confidence of pseudo data (label smoothing for pseudo text)
% data augmentation (harsher SpecAugment like noisy student learning scheme)

\section{TOOLKIT CHARACTERISTICS}
\label{sec:toolkit}
While the base model training follows the ordinary supervised training scheme as done in many toolkits, the last two stages need more specific designs to improve pipeline efficiency.
Our toolkit implements the following unique functionalities to simplify the pipeline of the TTS$\to$ASR chain.

\subsection{Fast model inference with uniform data structure}
\begin{figure}[t]
  \centering
  \includegraphics[width=\linewidth]{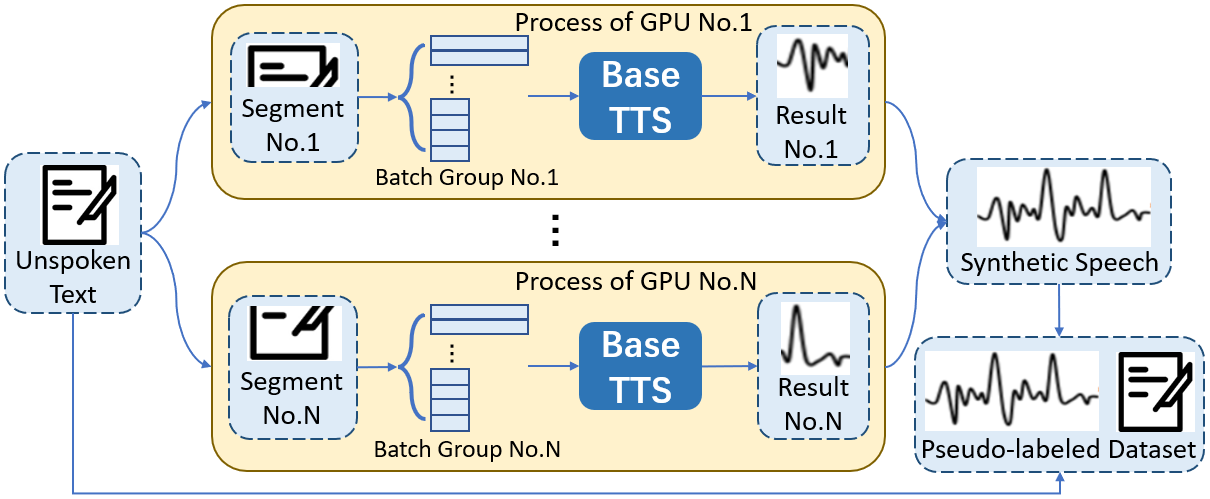}
  \caption{The flowchart of the batch-level TTS inference by multiple GPUs. 
%   In each process, CPUs fetch the data samples from the input dataset segment into a series of batches and send the processed batches to the GPU for parallel model calculation. After GPU calculation, CPUs collect the results and organize them into savable formats. 
  We group unlabeled text by the summation of their lengths to ensure the same number of synthetic frames in each batch. ASR beam searching shares a similar mechanism.}
  \label{fig:model_inference}
\end{figure}

Due to the auto-regressive nature of E2E ASR\&TTS models, model inference consumes plenty of time for large amounts of unlabeled data in the second stage. 
Our toolkit implements batch-level model inference by \emph{torch.nn.parallel.Distributed-DataParallel} that simultaneously utilizes multiple GPUs to speed up model inference. 
During inference, we partition unlabeled data into multiple non-overlapping segments. Each GPU holds an exclusive process that decodes a batch of unlabeled data in its responsible part at one inference step. 
After all the GPUs finish their assignments, we combine all the inference results to form the final pseudo-labeled dataset.
Fig \ref{fig:model_inference} shows an illustration of TTS decoding by multiple GPUs.

During model inference, we decouple the model calculation from the data storage so that the model hypotheses and inference metadata (e.g., WER, ASR confidence) could have the same structure as the real datasets. 
% 可删
The uniform data structure for real and pseudo datasets dramatically reduces the effort required for us to conduct the experiments.

\subsection{Batch generation by multiple dataloaders}
\label{sec:multiple_dataloader}
When training the target ASR model in the final stage, it's not a good idea to mix the real-labeled dataset with the pseudo-labeled one and dynamically fetch data from the mixed one. In most cases, unlabeled data is way more than labeled data, which results in plenty of data batches dominated by pseudo-labeled data. 
Therefore, it's hard to control the gradient direction of each data batch, and training becomes unstable. 

\begin{figure}[t]
  \centering
  \includegraphics[width=\linewidth]{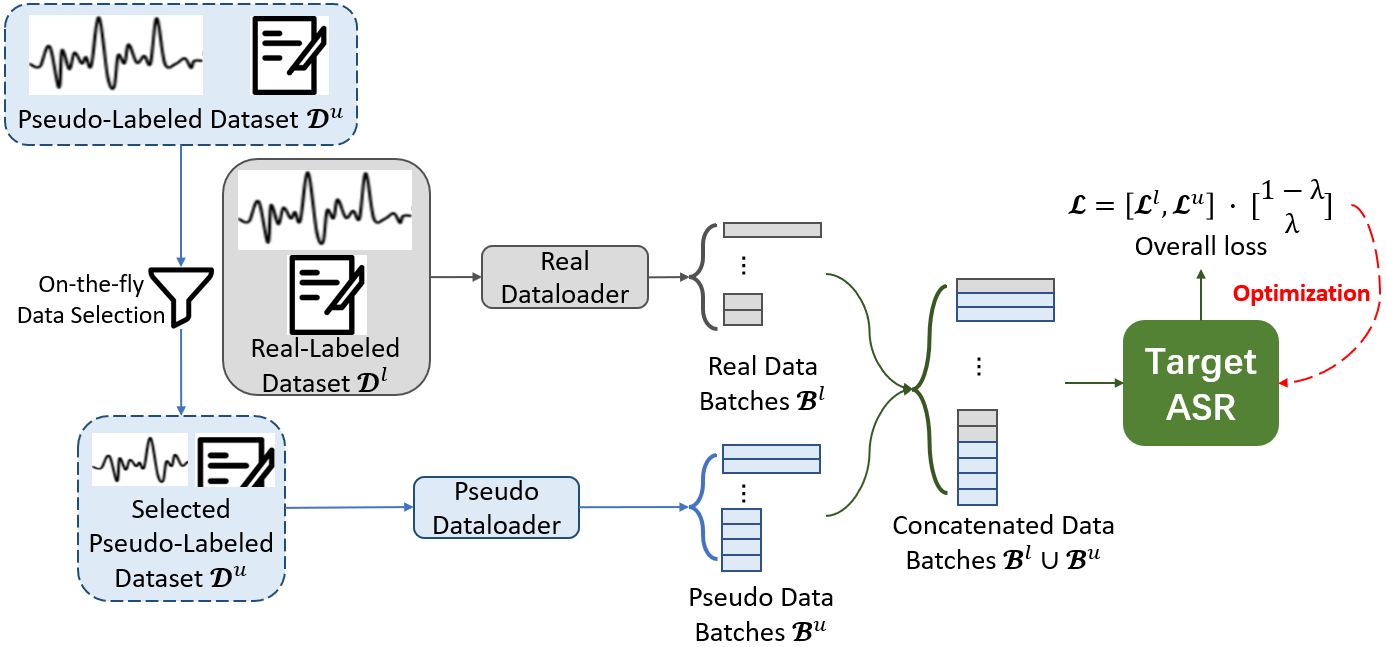}
  \caption{The flowchart of data loading by two dataloaders with on-the-fly data selection. 
  We concatenate real and pseudo batches with similar utterance lengths to reduce the number of model forwarding operations.}
  \label{fig:data_loading}
\end{figure}

Our toolkit utilizes multiple \emph{torch.utils.data.DataLoader} to separately fetch speech-text pairs from different datasets, which generates the batches with a static real-pseudo data ratio, as shown in Fig. \ref{fig:data_loading}.
The static data ratio not only regularizes the gradient direction calculated by each batch but also makes an evident real-pseudo composition of each batch. 
The evident composition allows us to design more advanced algorithms for better usage of pseudo-labeled data during training.

% single-dataloader: 
%   Pros: low CPU burden, fewer workers to fetch data from the disk
%   Cons: unbalanced batch composition, no separate processing & weight adjustment for real & pseudo data
% multi-dataloader:
%   Pros: High CPU burden, more workers to fetch data from the disk
%   Cons: balanced batch composition, separate processing & weight adjustment for real & pseudo data

\subsection{On-the-fly data selection for pseudo-labeled data}
% torch.utils.data.Dataset
Our toolkit decouples the batch generation from the dataset access (i.e., \emph{
torch.utils.data.Dataset}) so that we can easily change the accessible speech-text pairs of each dataloader by the given metadata during training. 
% distribution还可以帮助我们去选择pseudo data filtering threshold
We also provide metadata distribution histograms to help users choose the proper thresholds to filter out the low-quality pseudo-labeled data. 

\section{EXPERIMENTS}
\label{sec:exp}

\subsection{ASR baseline and topline of supervised learning}
% 介绍使用的数据集
% ASR模型结构，数据前端处理方法，优化方法
% train_clean_100 base model, noam lr scheduler, small learning rates for stable training, not able to converge (early overfitting) with normal warmup plus square root decay scheduler
Our ASR models are based on Speech-Transformer \cite{dong2018speech}, where the encoder and decoder are composed of 12 and 6 Transformer layers, respectively. 
We make two model configurations for ASR baseline ($d_{model}$=256, $n_{head}$=4) and topline ($d_{model}$=512, $n_{head}$=8). 
The two configurations share the same 2048-dimensional feed-forward layers. 
We uniformly do ASR decoding by 16 beams without a CTC branch for all experiments. 
% We adjust the softmax temperature used during beam searching based on the WER performance on \emph{dev\_clean} for each experiment. 
The results of our baseline and topline models are shown in Table \ref{tab:asr_base_top}.

% 为了空间可以删除
% The number of batches per epoch is set to 1000 for both baseline and topline training.
% We use the Adam optimizer \cite{kingma2014adam} and schedule the learning rates by a linear warm-up and square root decay where the number of warm-up epochs is set to 32 for baseline and 16 for topline. The peak value of learning rates after warm-up is set to 0.001. 
% 这段话有需要可以删除
% To further regularize ASR training, we introduce a layer normalization layer \cite{chen2020multispeech} for the encoder and decoder embedding vectors before adding them to the positional encoding. 
% Also, we found it helpful to add a learnable scalar \cite{li2019neural} for the positional encoding which enables us to train our baseline models with larger learning rates. 

\begin{table}[]
    \centering
    \resizebox{\columnwidth}{!}{
    \begin{tabular}{|l|c|c|c|}
        \hline
        Work (Model) & dev\_clean & test\_clean & test\_other\\
        \hline
        \multicolumn{4}{|c|}{Baseline on LibriSpeech-\emph{train\_clean\_100}}\\
        \hline
        \cite{rosenberg2019speech} (LAS \cite{chan2016listen}) & - & 12.46\% & 34.00\%\\
        % \hline
        \cite{kahn2020self} (TDS \cite{hannun2019sequence}) & 14.00\% & 14.85\% & 39.95\%\\
        % \hline
        \cite{higuchi2022momentum} (Transformer \cite{dong2018speech}) & 12.20\% & 12.90\% & 31.10\%\\
        % \hline
        Ours (Transformer) & \textbf{11.90\%} & \textbf{12.20\%} & \textbf{30.19\%}\\
        \hline
        \multicolumn{4}{|c|}{Topline on}\\
        \multicolumn{4}{|c|}{LibriSpeech-\emph{train\_clean\_100} \& LibriSpeech-\emph{train\_clean\_360}}\\
        \hline
        \cite{rosenberg2019speech} (LAS) & - & 6.30\% & 22.41\%\\
        % \hline
        \cite{yue2021exploring} (LAS) & - & 6.50\% & -\\
        % \hline
        Ours (Transformer) & 5.08\% & \textbf{5.65\%} & \textbf{15.91\%}\\
        \hline
        \multicolumn{4}{|c|}{Topline on}\\
        \multicolumn{4}{|c|}{LibriSpeech-\emph{train\_clean\_100} \& LibriTTS-\emph{train\_clean\_360}}\\
        \hline
        Ours (Transformer) & 6.14\% & 6.57\% & 19.50\%\\
        \hline
    \end{tabular}}
    \caption{Word error rate (WER) of ASR baseline and topline without the language model compared to the literature. For the baseline, we use the vocabulary with 1k BPE tokens; for the topline, we use the vocabulary with 5k BPE tokens. 
    We trained two topline models on LibriSpeech and LibriTTS to make a fair comparison with our TTS$\to$ASR chain.}
    \label{tab:asr_base_top}
\end{table}

\subsection{Base TTS model training}
As mentioned in \cite{rossenbach2020generating}, LibriSpeech \cite{panayotov2015librispeech} is unsuitable for training TTS models, so we downsampled LibriTTS \cite{zen2019libritts} to 16kHz for both base TTS training (\emph{train\_clean\_100}) and speech synthesis (\emph{train\_clean\_360}).
Our TTS model is based on MultiSpeech \cite{chen2020multispeech}, where the encoder and decoder are composed of the same 6 Transformer layers ($d_{model}$=512, $n_{head}$=8, $d_{ff}$=2048). The encoder directly consumes characters as the input.

We observed that TTS is data-hungrier than ASR. 
Therefore, we turn all the lowercase letters into capital versions and remove punctuation marks other than commas, periods, and single quotes to help TTS convergence. 
We make our base TTS model generate four consecutive frames at each decoding step to speed up the speech synthesis. 
We don't adopt the neural vocoder and use the same 80-dimensional log-Mel spectral features extracted by 50ms window length and 10ms window shift for both ASR and TTS training. 
Also, we found it helpful for base TTS training if we scale up the encoder and decoder embedding vectors by $\sqrt{d_{model}}$ before adding them to the positional encodings. 
During TTS speech synthesis, we randomly sample speakers from LibriTTS-\emph{train\_clean\_100} as reference speakers for our base TTS model.

\begin{table}[]
    \centering
    \resizebox{\columnwidth}{!}{
    \begin{tabular}{|c|c|c|c|}
        \hline
        Filtering Thresholds & dev\_clean & test\_clean & test\_other\\
        \hline
        \multicolumn{4}{|c|}{TTS$\to$ASR Chain by LibriTTS-\emph{train\_clean\_360}}\\
        \hline
        Base WER$<$10\% & 9.27\% & 10.22\% & 27.12\%\\
        % \hline
        Base WER$<$20\% & 8.48\% & 9.30\% & 25.90\%\\
        % \hline
        Base WER$<$30\% & 8.11\% & 8.68\% & 24.82\%\\
        % \hline
        Base WER$<$50\% & 7.83\% & 8.76\% & 25.65\%\\
        % \hline
        Base WER$<$100\% & 7.81\% & 8.45\% & 24.50\%\\
        % \hline
        No Base WER Filtering & \textbf{7.32\%} & \textbf{8.07\%} & \textbf{24.17\%}\\
        \hline
        \multicolumn{4}{|c|}{ASR Self-Training by LibriSpeech-\emph{train\_clean\_360}}\\
        \hline
        Top20\% Confidence & 10.23\% & 10.80\% & 26.28\%\\
        % \hline
        Top40\% Confidence & 9.04\% & 9.75\% & 24.26\%\\
        % \hline
        Top60\% Confidence & 8.86\% & 9.47\% & 23.44\%\\
        % \hline
        Top80\% Confidence & \textbf{8.51\%} & 9.36\% & 22.76\%\\
        % \hline
        No Confidence Filtering & 8.67\% & \textbf{9.33\%} & \textbf{22.66\%}\\
        \hline
    \end{tabular}}
    \caption{WER of target ASR models with different filtering thresholds.}
    \label{tab:asr_wer_filter}
\end{table}

\subsection{WER filtering by the base ASR model}
Our target ASR models have the same configuration as our topline model. 
We vary the base WER filtering thresholds as shown in Table \ref{tab:asr_wer_filter}, where $\lambda$ is uniformly set to 0.5. 
% Since the WER performance of the base ASR model is more than 10\% on the clean real speech, we set 10\% as the lowest filtering threshold. 
% For stable training, we also filter out 10\% of synthetic speech with the largest frame-to-char ratio which is the quotient of the number of log-Mel frames in the synthetic speech over by the number of characters in the unspoken text. 
We compare our TTS$\to$ASR chain with ASR self-training \cite{kahn2020self}, where the base ASR model generates pseudo text for the untranscribed speech to enlarge the training set on its own. 
In the experiments of ASR self-training, we used LibriSpeech-\emph{train\_clean\_360} as the unlabeled data and the ASR decoding confidence as the filtering criterion.

% 分析实验结果
% 无WER过滤的效果最好，这也就意味着合成语音即使存在一些错误，也是对ASR训练有好处的，类似于一种数据增强？
The results show that the base WER filtering instead reduces the ASR improvement. 
We found that the WERs given by the base ASR failed to evaluate the quality of synthetic speech because of the mismatch between real speech and synthetic speech.
% We found that the base WERs larger than 100\% are mostly caused by repeating phrases while the synthetic speech with lower WERs contains more silence and mispronunciations. 
However, the target ASR still benefits from training on synthetic speech even though there are some errors in them. 
We hypothesize that some errors in the synthetic speech may act as data augmentations, e.g., silence could force ASR to predict the missing parts.

% other上的提升并没有clean的大，因为TTS是在clean上训练的，所以合成语音还是属于clean domain，other和clean之间可能存在着acoustic mismatch，因此对other的提升没有很大。
We also observed that the improvement of the TTS$\to$ASR chain on \emph{test\_other} is smaller than ASR self-training. 
We hypothesize that there is an acoustic mismatch of speech data between \emph{clean} and \emph{other} datasets.
Since our TTS model is trained only on \emph{clean} data, the synthetic speech fails to provide the target ASR model with much acoustic details as the real untranscribed speech does.

\subsection{The effect of real-synthetic data composition}
% single-dataloader from the mixed dataset of the real-labeled and pseudo-labeled data
To study the influence of batch compositions on ASR training mentioned in Sec \ref{sec:multiple_dataloader}, we conducted a contrast experiment in different batch generations shown in Table \ref{tab:asr_data_balance}. 
We considered three scenarios where the training set contains 100\%, 66\%, and 33\% of real speech. 
In each scenario, the batch compositions are either static by two dataloaders or dynamic by one dataloader.
Our results indicate that as long as synthetic speech exists in the training set, it's necessary to include a certain amount of real speech in each batch to regularize the gradient directions.
But for topline training on two real datasets, it's better to mix them so that the model training could enjoy more randomness on the batch generation.

\begin{table}[]
    \centering
    \begin{tabular}{|c|c|c|c|}
        \hline
        Batch Composition & dev\_clean & test\_clean & test\_other\\
        \hline
        \multicolumn{4}{|c|}{Topline on}\\
        \multicolumn{4}{|c|}{LibriSpeech-\emph{train\_clean\_100} \& LibriTTS-\emph{train\_clean\_360}}\\
        \hline
        Static R2S Ratio & 6.37\% & 6.82\% & 19.75\%\\
        % \hline
        Dynamic R2S Ratio & \textbf{6.14\%} & \textbf{6.57\%} & \textbf{19.50\%}\\
        \hline
        \multicolumn{4}{|c|}{10\% Base WER Filtering (2:1)}\\
        \hline
        Static R2S Ratio & \textbf{9.27\%} & \textbf{10.22\%} & \textbf{27.12\%}\\
        % \hline
        Dynamic R2S Ratio & 10.59\% & 11.39\% & 28.97\%\\
        \hline
        % \multicolumn{4}{|c|}{30\% Base WER Threshold (1:1)}\\
        % \hline
        % Static & 8.11\% & 8.68\% & 24.82\%\\
        % \hline
        % Dynamic & \% & \% & \%\\
        % \hline
        \multicolumn{4}{|c|}{No Base WER Filtering (1:2)}\\
        \hline
        Static R2S Ratio & \textbf{7.32\%} & \textbf{8.07\%} & \textbf{24.17\%}\\
        % \hline
        Dynamic R2S Ratio & 7.75\% & 8.54\% & 24.93\%\\
        \hline
    \end{tabular}
    \caption{
    WER of target ASR models with different batch compositions. 
    R2S Ratio stands for the real-to-synthetic ratio of each training batch. 
    The numbers in the brackets indicate the R2S ratio of the entire training set.}
    \label{tab:asr_data_balance}
\end{table}

\section{CONCLUSION AND FUTURE WORK}
\label{sec:conclusion}
This paper introduces SpeeChain, a novel toolkit designed for the large-scale machine speech chain. 
This first version provides an easy-to-use pipeline of TTS data augmentation for semi-supervised ASR. 
Our extensive experiments demonstrate that semi-supervised ASR models carefully developed via the TTS$\to$ASR chain could achieve performance close to the supervised one trained on real-labeled data.
In our future work, we will implement various advanced ASR and TTS models and develop additional applications of the machine speech chain, such as the ASR$\to$TTS chain for semi-supervised TTS and online speech chain learning for joint ASR-TTS optimization. 

\section{Acknowledgements}
Part of this work is supported by JSPS KAKENHI (grant number JP21H05054 and JP21H03467) and Google AI grant.

% Our future work may consider to replace the log-Mel spectrogram with MFCC to not only focus on the phonetic information but also speed up TTS decoding.
% log-Mel Spectral features still contain too much speaker variability, hard for TTS to learn with a little amount of labeled data. 
% Try MFCC to remove the speaker variability and better capture phonetic information

% autoregressive Transformer-TTS takes a long time and a large amount of GPU memory for decoding, hard to be applied to large-scale semi-supervised learning scenario. 
% Try non-autoregressive TTS model like FastSpeech2

% offline semi-supervised learning we need to store all the pseudo data and filtering criteria to the disk, which also makes it less applicable to large-scale scenario. We consider to implement online semi-supervised ASR pipeline in the future where all the pseudo data and criteria will only produced and used in the RAM.

\footnotesize
\bibliographystyle{IEEEbib}
\bibliography{refs}

\end{document}